\documentclass[letterpaper]{article} % DO NOT CHANGE THIS
\usepackage{aaai25}  % DO NOT CHANGE THIS
\usepackage{times}  % DO NOT CHANGE THIS
\usepackage{helvet}  % DO NOT CHANGE THIS
\usepackage{courier}  % DO NOT CHANGE THIS
\usepackage[hyphens]{url}  % DO NOT CHANGE THIS
\usepackage{graphicx} % DO NOT CHANGE THIS
\usepackage{natbib}  % DO NOT CHANGE THIS AND DO NOT ADD ANY OPTIONS TO IT
\usepackage{caption} % DO NOT CHANGE THIS AND DO NOT ADD ANY OPTIONS TO IT
\usepackage{algorithm}
\usepackage{algorithmic}
\usepackage{amsmath}  % 或者使用 amssymb
\usepackage{amssymb}
\usepackage{ragged2e}
\usepackage{multirow}
\usepackage[cmyk]{xcolor}
\usepackage{colortbl}
\usepackage{color}
\usepackage{arydshln}
\usepackage{cleveref}
\usepackage{newfloat}
\usepackage{listings}
\usepackage{subfigure}
\usepackage{enumerate}

\DeclareCaptionStyle{ruled}{labelfont=normalfont,labelsep=colon,strut=off} % DO NOT CHANGE THIS
\floatstyle{ruled}
\newfloat{listing}{tb}{lst}{}
\floatname{listing}{Listing}
\title{Robust Self-Paced Hashing for Cross-Modal Retrieval with Noisy Labels}
\author{
    Ruitao Pu \textsuperscript{1}, 
    Yuan Sun \textsuperscript{1}\thanks{Corresponding author.},
    Yang Qin \textsuperscript{1},
    Zhenwen Ren \textsuperscript{2},
    Xiaomin Song \textsuperscript{3},
    Huiming Zheng \textsuperscript{3}, \\
    Dezhong Peng \textsuperscript{1,3}
}
\affiliations{
    \textsuperscript{\rm 1}Sichuan University, Chengdu, China, 610044,\\
    \textsuperscript{\rm 2}Southwest University of Science and Technology, Mianyang, China, 621010,\\
    \textsuperscript{\rm 3}Sichuan National Innovation New Vision UHD Video Technology Co., Ltd., Chengdu, China, 610095,\\
    ruitaopu@gmail.com, sunyuan\_work@163.com, qinyang.gm@gmail.com, rzw@njust.edu.cn, songxiaomin@uptcsc.com, michaelzheng@uptcsc.com, pengdz@scu.edu.cn.
    
}

\begin{document}
\sloppy
\maketitle

\begin{abstract}
Cross-modal hashing (CMH) has appeared as a popular technique for cross-modal retrieval due to its low storage cost and high computational efficiency in large-scale data. Most existing methods implicitly assume that multi-modal data is correctly labeled, which is expensive and even unattainable due to the inevitable imperfect annotations (i.e., noisy labels) in real-world scenarios. Inspired by human cognitive learning, a few methods introduce self-paced learning (SPL) to gradually train the model from easy to hard samples, which is often used to mitigate the effects of feature noise or outliers. It is a less-touched problem that how to utilize SPL to alleviate the misleading of noisy labels on the hash model. To tackle this problem, we propose a new cognitive cross-modal retrieval method called Robust Self-paced Hashing with Noisy Labels (RSHNL), which can mimic the human cognitive process to identify the noise while embracing robustness against noisy labels. Specifically, we first propose a contrastive hashing learning (CHL) scheme to improve multi-modal consistency, thereby reducing the inherent semantic gap. Afterward, we propose center aggregation learning (CAL) to mitigate the intra-class variations. Finally, we propose Noise-tolerance Self-paced Hashing (NSH) that dynamically estimates the learning difficulty for each instance and distinguishes noisy labels through the difficulty level. For all estimated clean pairs, we further adopt a self-paced regularizer to gradually learn hash codes from easy to hard. Extensive experiments demonstrate that the proposed RSHNL performs remarkably well over the state-of-the-art CMH methods.

% CMH aims to learn discriminative binary codes to alleviate the cross-modal gap. However, the performance of existing CMH methods heavily relies on well-annotated data, while low-quality annotations are ubiquitous due to time and labor constraints. Although some CMH methods are proposed to learn hash codes with noisy labels, most of them treat every training instance equally, which may cause the model to be biased toward hard instances with noisy labels and result in the model overfitting.

\end{abstract}
% NSH presents a dynamic hardness measurement strategy that dynamically estimates the learning difficulty of each pair and distinguishes the noisy labels while facilitating learning to hash from easy to hard.

% NSH presents a dynamic hardness measurement strategy that dynamically estimates the learning difficulty of each pair and distinguishes the noisy labels while facilitating learning to hash from easy to hard.

% assigns lower weights for the hard pairs

%  automatically assesses the learning order usually based on the feedback of the hashing model.

% learn efficient and robust hash representations

\begin{links}
    \link{Code}{https://github.com/perquisite/RSHNL}
\end{links}
% \begin{links}
% \link{Code}{{https://github.com/perquisite/RSHNL}
% \end{links}

\section{Introduction}
With the explosive growth of multi-modal data, cross-modal retrieval (CMR) has attracted a wide range of attention in the community \cite{zhou2023state}, which retrieves relevant samples across different modalities.
% which mainly uses a query in one modality to retrieve the relevant samples in another modality. 
For large-scale multi-modal data, cross-modal hashing (CMH) \cite{zhu2023multi} offers an efficient solution due to its low storage cost and high retrieval efficiency. The basic idea of CMH is to learn discriminative hash codes to alleviate the heterogeneity gap between different modalities. Due to the complexity of collecting annotations, some unsupervised CMR methods \cite{zhang2023semi, li2024romo} have been proposed to eliminate the reliance on abundant labels. However, their performance often suffers without supervised semantic guidance. Recently, numerous supervised CMH methods \cite{li2024cross, chen2021deep} have been proposed and achieved pleasing performance. Most of them implicitly assume that all collected labels are correctly labeled, which is unrealistic due to inevitable noisy labels from manual or non-expert annotations \cite{song2022learning, kuznetsova2020open}. These noisy labels can mislead hash models, significantly degrading retrieval performance.
% . However, in real-world scenarios, such an assumption is impractical, since manual annotation \cite{song2022learning} or non-expert annotations \cite{kuznetsova2020open} inevitably bring some noisy labels.
% Hence, this will easily mislead hash models, thereby leading to a significant degradation in retrieval performance.
Besides, learning from multi-modal instances with noisy labels is difficult due to a great heterogeneity gap. Thus, it is a challenging problem to mitigate the performance degradation caused by noisy labels for cross-modal retrieval.   

% MRL \cite{MRL} proposes a robust cluster mechanism to amplify the loss from clean samples, thereby resisting noise effects. But MRL could cause the cluster centers to drift. To stabilize cluster centers, 
% which is applied to asymmetric noise label scenarios
% Defining a full learning order based on the difficulty of sample features with noisy labels is extremely challenging. 

% Afterward, WASH \cite{WASH} estimates noisy labels to perform enhanced semantic-aware hash learning. 
To alleviate the impact of noisy labels, many CMR methods \cite{ELRCMR, wang2024robust, NrDCMH} have been developed. For example, ELRCMR utilizes dynamic weights to prevent overfitting noisy labels. However, most of them rely on real-valued representation, leading to high storage and computational costs. Although hash representations are more lightweight, unreliable labels could expand quantization errors. To this end, a few CMH methods have been proposed. For instance, NrDCMH adopts the difference between label similarity and feature similarity to detect noise. Although remarkable progress, most of them implicitly assume constant learning priorities for each instance, making the model biased toward hard instances with noisy labels and leading to the overfitting problem. Inspired by human cognitive learning, self-paced learning (SPL) was presented to explore more valuable discriminative information contained in hard instances gradually. In other words, SPL can gradually train the model from easy to hard instances to improve generalization. However, SPL is usually used for feature noise or outliers. It is a less-touched problem to employ the SPL paradigm to mitigate the negative effects of noisy labels.

\begin{figure*}[ht]
\centering
\setlength{\abovecaptionskip}{0.1cm}
\includegraphics[width=0.75\textwidth]{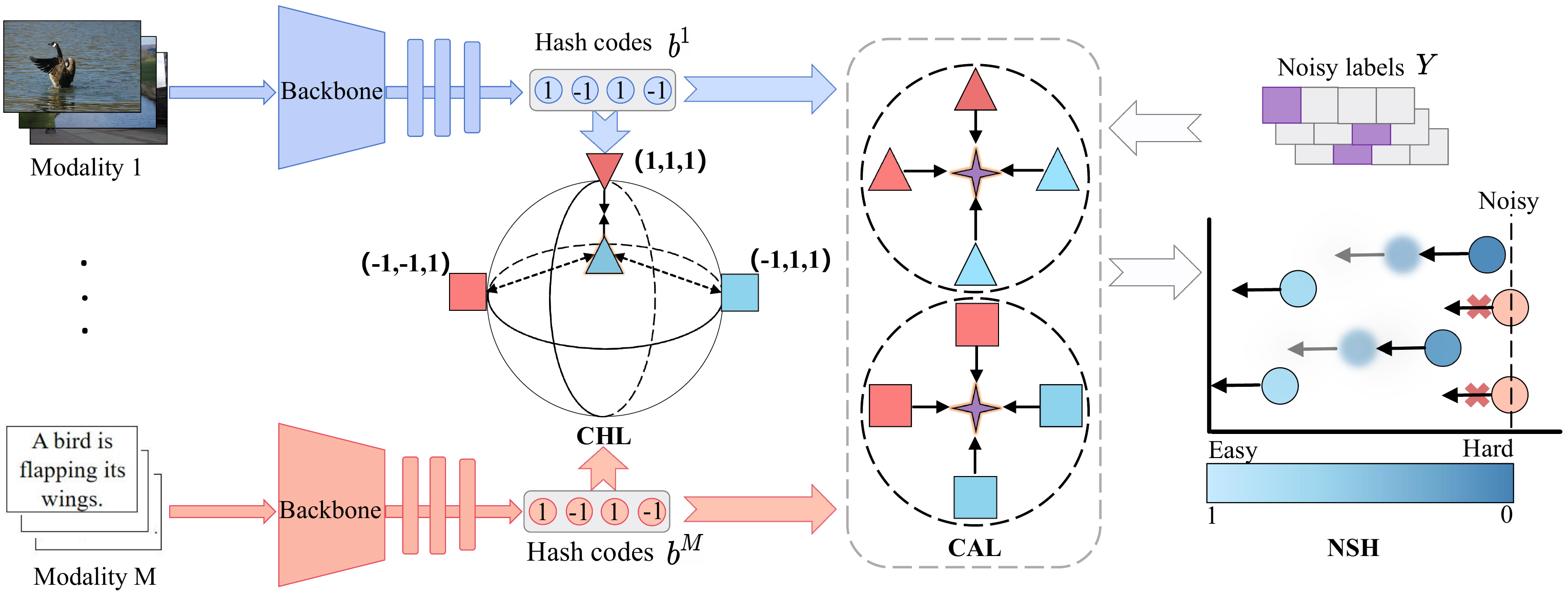}
\caption{The framework of our RSHNL. Blue and red represent hash codes of different modalities. Triangles and rectangles represent different categories. And the doji represents hash centers. Specifically, CHL maximizes the consistency of multi-modal data to alleviate the cross-modal gap. CAL develops a unified hash code for each class as a center and encourages the compactness of intra-class hash codes towards their corresponding hash centers. NSH dynamically distinguishes noisy labels based on their assessment difficulty while facilitating learning hash codes from easy to hard for clean pairs.}
\label{fig:1}
\end{figure*}
% present a dynamic hardness measurement strategy to estimate the learning difficulty of each instance with noisy labels. 
% the prioritizing of instances’ learning orders
% The learning priority remains the same 
% To smoothly learn the hash model from easy instances to hard ones, we design a self-paced regularizer to adaptively measure the instances’ learning hardness
% smoothly update the model from easier data to more difficult ones.
% The proposed NSH enables the model to learn from noisy labels, progressing from easier to harder instances.
% gradually focus on hard pairs 
% NSH dynamically estimates the difficulty of each pair and distinguishes the noisy labels while facilitating learning to hash from easy to hard.
%can enable our model to eliminate noisy labels 
% beginning with easier pairs and advancing to more complex ones. 
In this paper, we propose a new cognitive cross-modal retrieval method, termed Robust Self-paced Hashing with Noisy Labels (RSHNL), which could enable the model to learn with noisy labels. As shown in Fig.\ref{fig:1}, our RSHNL mimics the human cognitive process to learn each instance in Hamming space from easy to hard, thereby embracing the robustness of eliminating noisy labels. Compared with existing self-paced hashing methods which only consider feature noise/outliers and cannot deal with noisy labels, our RSHNL assesses the learning pace by identifying noisy labels and learning clean pairs from easy to hard in the form of cognitive learning. Specifically, to reduce the inherent semantic gap, we first present a contrastive hashing learning (CHL) scheme to maximize the consistency between multi-modal data. Then, we propose center aggregation learning (CAL) to learn unified hash centers to aggregate hash codes from the same category, thereby mitigating the intra-class variations from multi-modal inputs. Further, we propose Noise-tolerance Self-paced Hashing (NSH) to adaptively measure the learning hardness of each instance and distinguish noisy labels according to the corresponding difficulty. For all sample pairs with clean labels, we adopt a self-paced regularizer that begins with easier pairs and advances to more complex ones. The main contributions are summarized as follows:
\begin{itemize}
\item We propose a new cognitive cross-modal hashing paradigm that alleviates the negative effect of noisy labels. To the best of our knowledge, this could be the first work that introduces SPL to distinguish and eliminate noisy labels in CMH tasks. 
\item We propose a Noise-tolerance Self-paced Hashing (NSH) loss that automatically determines noise labels and builds a full learning sequence for clean sample pairs, thereby evolving from easy to hard until all clean pairs are incorporated for training.
\item Extensive experiments comprehensively verify that our proposed RSHNL has remarkably superior performance over the current state-of-the-art methods in different noise rates.
\end{itemize}

% RSHNL consists of a self-paced robust aggregation learning (SRA) and a self-supervised contrastive learning (SC) mechanism. Specifically, SRA is proposed to estimate whether the labels are noisy by measuring the difficulty of the training instances. Then, clean instances will be selected for training and the model will be trained from simple samples to hard samples, embracing more robustness and generalization of the model simultaneously. SC is proposed to learn discriminative hash codes by extracting instance-level contrastive information and mitigate the modality heterogeneity gap by maximizing instance co-occurrence information. 

% To the best of our knowledge, this may be one of the first studies to use SPL to solve learning with noisy labels for cross-modal hashing.
% RSHNL can resist the negative impact of noisy labels and indent the cross-modal gap to learn discriminative hash codes with noisy labels.
% At the beginning of the training process, all instances are assigned equal weights to facilitate the deep hash model in learning universal patterns.  Subsequently, higher weights are given to hard instances according to their hardness to capture more valuable discriminative information from them. 

\section{Related Work}
\subsection{Cross-modal Hashing}
Recently, many CMH methods have been proposed, which can be roughly divided into two categories, i.e., unsupervised and supervised. Unsupervised CMH methods \cite{cao2023generative} aim to utilize the original distribution of the data to learn a common Hamming space. For example, DSAH \cite{DSAH} integrates co-occurrence information and semantic relevance of different modalities to guide hashing learning. For supervised CMH methods \cite{liu2024dual, sun2024distribution}, their goal is to leverage annotation information to learn more compact and discriminative hash codes. For instance, DCMH \cite{DCMH} maintains the semantic relevance of hash representations through the label similarity matrix. 
% However, existing CMH methods usually have two limitations: solution space compression and loss function oscillation. To overcome the above problems, SCH \cite{SCH} has been proposed, which deals with instances with different similarities distinctively and adopts the semantic channel. 
However, almost all of these methods implicitly assume multi-modal data are well labeled. In practical applications, labeling noise is ubiquitous, which could mislead the hash model to overfit the noise. To tackle this problem, some supervised methods are developed to learn to hash from noisy labels robustly. For example, to utilize the memorization effect of DNNs and reduce the impact of noisy labels, CMMQ \cite{CMMQ} selects confidence samples with smaller loss values. To deal with the problem of low-quality label annotations, DHRL \cite{DHRL} constructs a ranking and swapping module to estimate the uncertainty from noisy labels. Although these methods achieve promising performance, they unconsciously ignore the adverse impact of sample pairs with noisy labels and keep the learning priority of each instance the same, which violates the human cognitive process.

% To explore the underlying manifold structure of multi-modal data, UGACH \cite{UGACH} adopts generative adversarial networks to improve the ability of unsupervised representation learning. 
% NrDCMH \cite{NrDCMH} detects noise by the differences between label similarity and feature similarity, and trains sample pairs. 

\subsection{Self-paced Learning}
Inspired by human cognitive learning, self-paced learning (SPL) \cite{2010-SPL-1, 2014-SPL-2} is proposed to train the model from easy samples to hard ones. For example, to mitigate the noise/outlier problem, SCSM \cite{liang2016self} utilizes SPL to learn samples from easy to hard. However, this will introduce several hyperparameters due to manually setting the weighting function, which is undesirable. Thus, Meta-SPN \cite{wei2021meta} is proposed to learn the weight values automatically. However, most of these methods only focus on the contributions of instances to learn hash codes from easy to hard, thereby resisting feature noise interference. To this end, DSCMH \cite{DSCMH} proposes a novel dual SPL mechanism from the perspectives of instance-level and feature-level difficulty to enhance robustness. In contrast, DHaPH \cite{DHaPH} pays more attention to difficult sample pairs, and assigns larger weights to the hard pairs to learn discriminative information. Although these SPL methods have achieved considerable performance, almost all of them only consider the noise or outliers in the data. When faced with multi-modal data with noisy labels, how to utilize SPL to alleviate the noise overfitting problem is rarely studied. In this paper, we expect to distinguish sample pairs with noisy labels and gradually explore discriminative semantic information from clean data for the hash model, thereby alleviating the negative effect of noisy labels.

% We denote the $M$-category modalities dataset with $N$ instances as $\mathcal{D} =\left \{ \mathcal{D} _m \right \} _{m=1}^M$, where ${D}_m =\left \{ (x_i^m,y_i^m) \right \} _{i=1}^N$, $x_i^m$ denotes the $i$-th sample of the $m$-th modality, $y_i^m\in\mathbb{R}^K$ denotes the label (may be noisy) of $x_i^m$, and $K$ is number of the categories. For sample $x_i^m$, if it belongs to $k$-th category, the $k$-th element of $y_i^m$ is 1, i.e. $y_{i,k}^m = 1$, otherwise $y_{i,k}^m = 0$. 

\section{The Proposed Method}
\subsection{Problem Formulation}
In this paper, some notations are provided for a clear presentation. We first denote ${D}_m =\left \{ (x_i^m,y_i) \right \} _{i=1}^N$ as multi-modal data with $N$ instances from $M$ modalities, where $x_i^m$ represents $i$-th sample of the $m$-th modality, $Y\in\mathbb{R}^{N \times K}$ is the corresponding labels, and $K$ is number of the categories. For a instance $x_i^m$, if it belongs to $k$-th category, the $k$-th element of $y_i$ is 1, i.e. $y_{i,k} = 1$, otherwise $y_{i,k} = 0$. 

The basic idea of cross-modal hashing is to project multi-modal data into a common Hamming space by adopting different hash functions. Let hash code of $i$-th sample from $m$-th modality is $b_i^m\in\left \{ -1,1 \right \} ^L$, where $L$ represents hash length. And each hash function can be denoted as $\mathcal{H}^m (\cdot,\Theta^m)$, where $\Theta^m$ is the corresponding learnable parameters. Subsequently, we can apply the $sign$ function to obtain hash representation, i.e., 
\begin{equation}
\begin{aligned}
b_i^m = {sign}(\mathcal{H}^m(x_i^m,\Theta^m)).
\end{aligned}
\label{eq:1}	
\end{equation}
Since binary optimization is a typical NP-hard problem \cite{DHaPH}, we adopt a $tanh$ function to learn binary-like codes during training.

\subsection{Contrastive Hashing Learning}
To narrow the inherent semantic gap between multi-modal data, we present a contrastive hashing learning scheme (CHL) that maximizes the consistency of hash codes from different modalities to improve the discrimination. Specifically, we treat the same instances from different modalities as positive pairs and then encourage them to be close while keeping negative pairs far away. First, we define the probability that $x_i^m$ belongs to the $i$-th instance as
\begin{equation}
\begin{aligned}
q(i\left | x_i^m\right .)=\frac{\sum_{j=1}^{M}e^{b_i^m(b_i^j)^\top /\ \tau} }{\sum_{j=1}^{M}\sum_{z=1}^{N}e^{b_i^m(b_z^j)^\top /\ \tau}},
\end{aligned}
\label{eq:8}	
\end{equation}
where $\tau$ is a temperature parameter. Then, the CHL loss $\mathcal L_{C}$ could be written as maximizing a joint probability $\prod_{i=1}^{N} \prod_{m=1}^{M}  p(i\left | x_i^m\right)$ of all samples, which is equivalent to minimizing the following formula, i.e.,
\begin{equation}
\begin{aligned}
\mathcal L_{C}&=\frac{1}{N} \sum_{i=1}^{N}\sum_{m=1}^{M} (1-r)\frac{1-(q(i\left | x_i^m\right .))^r}{r} \\ &+ r(1-q(i\left | x_i^m\right .)) .
\end{aligned}
\label{eq:9}	
\end{equation}
By minimizing Eq.\ref{eq:9}, positive pairs are forced to be compressed, while negative pairs are scattered in the Hamming space, thereby alleviating multi-modal discrepancy. 

\subsection{Noise-tolerance Self-paced Hashing}
To mitigate the intra-class variations of multi-modal data, we propose center aggregation learning (CAL) to learn a uniﬁed hash representation for each class as a center and promote modality-specific hash codes with the same category to be aggregated to the corresponding hash centers. Specifically, we randomly initialize and obtain hash centers $C =\left \{c_1,c_2,...,c_K \right \}$, where $c_K\in \mathbb{R}^{L \times 1}$ is the normalized and discretized binary vector for the $K$-th class. For any sample $x_i^m$, 
% to encourage all hash codes from the same class to approach the corresponding center, 
we first define its probability belonging to the $k$-th center as
\begin{equation}
\begin{aligned}
p(k\left | x_i^m\right) = \frac{e^{b_i^mc_k/\ \tau}}{\sum_{j=1}^{K}e^{b_i^mc_j/\ \tau} }, 
\end{aligned}
\label{eq:3}	
\end{equation}
where $\tau$ is a temperature parameter. Due to the lack of guidance information, the obtained probabilities could lead to prediction errors. Thus, to make the hash code inherit more semantic information, we define the semantic aggregation probability as 
\begin{equation}
\begin{aligned}
v_i^m = \sum_{k=1}^{K} y_{i,k} p(k\left | x_i^m\right).
\end{aligned}
\label{eq:4}	
\end{equation}
Afterward, we can obtain the following center aggregation loss $\mathcal L_{p}$, i.e.,
\begin{equation}
\begin{aligned}
\mathcal L_{p}=\frac{1}{N} \sum_{i=1}^{N}\sum_{m=1}^{M} (1-r)\frac{1-(v_i^m)^r}{r} + r(1-v_i^m),
\end{aligned}
\label{eq:5}	
\end{equation}
where $r\in (0,1] $ is a weight factor. However, due to the ubiquitous noisy labels in the data, such center aggregation loss is inevitably disrupted, thereby tending to overfit the corrupted labels. Since DNNs \cite{a_survey_2022} are robust in the early training stage, we use Eq.\ref{eq:5} to warm up the model.

Inspired by the great success of self-paced learning (SPL), we can organize the learning sequence of samples from easy to hard, thereby improving the retrieval performance. Thus, some SPL-based hashing methods \cite{DSCMH,sun2024dual} have been proposed to mitigate the negative effects of noise or outliers. However, all of them implicitly assume that the multi-modal data are labeled correctly while ignoring the existence of noisy labels. Moreover, they keep the learning priority of each sample with noisy labels constant, which could be unreasonable due to the labeled differences. To overcome this issue, we propose a Noise-tolerance Self-paced Hashing (NSH) strategy to learn hash codes from noisy labels. Similar to prior SPL methods, our proposed NSH gradually learns from easy pairs to difficult ones, thereby automatically incorporating more data into the training process. Different from them, we reveal that the SPL scheme can distinguish sample pairs with noisy labels to mitigate the overfitting problem. Specifically, our NSH adopts a hardness measurement strategy that dynamically estimates the learning difficulty of each pair and distinguishes the noisy labels. Then, NSH gradually learns hash codes from easy to hard until it is sufficient to handle hard ones. Thereupon, the problem could be formulated as follows
\begin{equation}
\begin{aligned}
\mathcal{L} _{S}&=\frac{1}{N} \sum_{i=1}^{N} w_i\underbrace{\sum_{m=1}^{M}(1-r)\frac{1-(v_i^m)^r}{r} +r(1-v_i^m)}_{\ell_i}      \\
&+\frac{1}{N} \sum_{i=1}^{N}\mathcal{R}(w_i,\gamma ),
\end{aligned}
\label{eq:6}	
\end{equation}
where $w_i\in \left [ 0,1 \right ]$ is the importance weight of $i$-th sample pairs that evaluates the reliability of label dimension. $\mathcal{R}(w_i,\gamma )$ is the self-paced regularizer controlled by the learning pace parameter $\gamma$, which could assign a weight $w_i$ to estimate the learning difficulty of each instance. We adopt a linear interpolation function \cite{jiang2014self} to construct the following self-paced regularizer $\mathcal{R}(w_i,\gamma )$, i.e.,
\begin{equation}
\begin{aligned}
\mathcal{R}(w_i,\gamma )=\gamma (\frac{1}{2}{w_i}^2-w_i ).
\end{aligned}
\label{eq:7}	
\end{equation}

% with $\gamma$ increases,

In brief, the weight can be regarded as the easiness of each sample pair with noisy labels. If the weight is higher, the instance could be viewed as easier. The loss decreases gradually with the learning process, thus enlarging the weight. When $\ell_i > \gamma$, we consider this sample pair could be mislabeled and assign the weight as zero to represent the hardest/noisy sample pair. When $\ell_i \leq \gamma$, NSH first considers reliable/easy pairs at the beginning and then gradually incorporates unreliable/hard ones into training. In other words, we learn hash codes in the way of human cognitive learning (i.e., from easy to hard) until more clean pairs are incorporated into the hash model.

% more reliable (i.e., smaller loss  $\ell_i$)
% dynamically weight the training pairs and gradually assign larger weights to the hard pairs with noisy labels, thus enabling the hashing model to obtain discriminative hash codes from hard pairs

 %To further learn discriminative hash representation with noisy labels, we adopt a self-supervised contrastive learning mechanism (SC) to extract Instance-level discrimination information. Different from previous methods \cite{DSCMR, MARS}, SC adopts a self-supervised learning mechanism to be immune to noisy labels. First, we denote the probability that $x_i^m$ belongs to the $i$-th instance as:
%  \begin{equation}
%     q(i\left | x_i^m\right .)=\frac{\sum_{j=1}^{M}e^{b_i^m(b_i^j)^\top /\ \tau} }{\sum_{j=1}^{M}\sum_{z=1}^{N}e^{b_i^m(b_z^j)^\top /\ \tau}} 
% \end{equation}
% Next, we force positive pairs (multi-modal samples of the same instance) to be close while keeping negative pairs (samples from different instances) far apart.

\begin{table*}[htb]
\centering
\setlength{\tabcolsep}{1mm}
\caption{The MAP scores with different bit lengths on the XMedia dataset under different noise rates.}

\begin{tabular}{l|c|c|c|c|c|c|c|c|c|c|c|c|c|c|c|c|c|c}
\hline
 &  & Noise & \multicolumn{4}{c|}{0.2} & \multicolumn{4}{c|}{0.4} & \multicolumn{4}{c|}{0.6} & \multicolumn{4}{c}{0.8} \\ \cline{3-19} 
 \multirow{-2}{*}{Task}& \multirow{-2}{*}{Method} & Ref. & 16 & 32 & 64 & 128 & 16 & 32 & 64 & 128 & 16 & 32 & 64 & 128 & 16 & 32 & 64 & 128 \\ \cline{1-19} 
 & DGCPN & AAAI'21 & 49.0 & 64.2 & 47.4 & 51.3 & 49.0 & 64.2 & 47.4 & 51.3 & 49.0 & 64.2 & 47.4 & 51.3 & 49.0 & 64.2 & 47.4 & 51.3 \\
 & CIRH & TKDE'22 & 72.5 & 78.0 & 79.3 & 83.8 & 72.5 & 78.0 & 79.3 & 83.8 & \underline{72.5} & \underline{78.0} & \underline{79.3} & \underline{83.8} & \underline{72.5} & \underline{78.0} & \underline{79.3} & \underline{83.8} \\
 & UCCH & TPAMI'23 & 37.6 & 50.2 & 67.9 & 82.7 & 37.6 & 50.2 & 67.9 & 82.7 & 37.6 & 50.2 & 67.9 & 82.7 & 37.6 & 50.2 & 67.9 & 82.7 \\  
 & WASH & TKDE'23 & 80.7 & 85.9 & 87.1 & 87.7 & 75.6 & 79.4 & 81.3 & 82.1 & 44.8 & 53.8 & 57.4 & 59.6 & 13.9 & 16.3 & 17.2 & 19.2  \\
 & HCCH & TMM'24 & 71.1 & 81.9 & 82.1 & 84.4 & 71.6 & 76.8 & 78.2 & 80.6 & 60.3 & 61.5 & 61.5 & 72.5 & 26.3 & 41.2 & 41.1 & 48.5  \\
 & DSCMH & AAAI'24 & 81.2 & 85.6 & \underline{87.9} & 88.2 & \underline{79.4} & \underline{83.6} & \underline{85.4} & 85.8 & 63.1 & 70.4 & 74.4 & 77.5 & 31.0 & 35.1 & 43.7 & 47.2  \\ 
 & HMAH & TMM'22 & 78.7 & 84.4 & 86.7 & 88.2 & 55.0 & 65.1 & 71.4 & 74.6 & 22.3 & 29.8 & 33.5 & 37.6 & 8.7 & 9.6 & 10.7 & 10.7 \\
 & CMMQ & CVPR'22 & \underline{86.6} & \underline{87.9} & 87.4 & 86.6 & 74.8 & 77.4 & 74.5 & 72.0 & 51.5 & 43.8 & 38.9 & 38.7 & 18.3 & 18.5 & 13.7 & 12.6 \\
 & MIAN & TKDE'23 & 18.1 & 29.6 & 34.5 & 35.4 & 12.4 & 16.2 & 17.7 & 16.2 & 10.7 & 9.9 & 11.5 & 11.1 & 7.6 & 7.7 & 7.1 & 8.0 \\
 & DHRL & TBD'24 & 11.0 & 39.3 & 86.7 & \underline{90.5} & 9.9 & 66.0 & 84.2 & \underline{86.6} & 9.4 & 37.7 & 69.6 & 74.4 & 6.6 & 8.9 & 37.6 &  43.8 \\ 
 & DHaPH & TKDE'24 & 79.3 & 84.9 & 86.9 & 88.6 & 69.6 & 78.2 & 82.6 & 84.6 & 52.3 & 66.0 & 72.8 & 79.8 & 42.7 & 48.3 & 60.6 & 70.2 \\
\multirow{-14}{*}{I2T} & RSHNL & Ours & \textbf{89.5} & \textbf{90.0} & \textbf{90.9} & \textbf{90.8} & \textbf{89.7} & \textbf{90.2} & \textbf{91.1} & \textbf{89.8} & \textbf{83.0} & \textbf{87.7} & \textbf{88.9} & \textbf{88.4} & \textbf{81.7} & \textbf{88.1} & \textbf{87.6} & \textbf{84.7} \\ \hline
 & DGCPN & AAAI'21 & 50.0 & 58.2 & 31.5 & 40.2 & 50.0 & 58.2 & 31.5 & 40.2 & 50.0 & 58.2 & 31.5 & 40.2 & 50.0 & 58.2 & 31.5 & 40.2 \\
 & CIRH & TKDE'22 & 67.4 & 73.1 & 76.9 & 82.0 & 67.4 & 73.1 & 76.9 & 82.0 & \underline{67.4} & \underline{73.1} & \underline{76.9} & 82.0 & \underline{67.4} & \underline{73.1} & \underline{76.9} & 82.0 \\
 & UCCH & TPAMI'23 & 56.0 & 66.9 & 75.5 & 83.8 & 56.0 & 66.9 & 75.5 & 83.8 & 56.0 & 66.9 & 75.5 & \underline{83.8} & 56.0 & 66.9 & 75.5 & \underline{83.8} \\
 & WASH & TKDE'23 
& 81.6 & \underline{86.0} & 86.9 & 88.3 & 75.0 & 78.7 & 81.3 & 82.1 & 44.6 & 53.6 & 56.7 & 59.4 & 14.3 & 16.7 & 17.3 & 19.4  \\
 & HCCH & TMM'24 & 69.5 & 80.7 & 80.0 & 84.0 & 70.0 & 75.7 & 76.7 & 80.0 & 58.1 & 60.2 & 58.5 & 72.6 & 26.2 & 40.8 & 41.2 & 48.2  \\
 & DSCMH & AAAI'24 & 80.5 & 85.0 & 86.3 & 87.9 & \underline{78.0} & \underline{81.6} & 83.5 & 84.8 & 63.5 & 70.9 & 73.9 & 77.4 & 29.1 & 33.0 & 41.0 & 45.0  \\
 & HMAH & TMM'22 & 77.4 & 84.0 & 85.9 & 88.0 & 53.2 & 65.0 & 71.0 & 73.9 & 23.1 & 30.5 & 34.1 & 38.9 & 8.9 & 10.0 & 11.5 & 11.4 \\
 & CMMQ & CVPR'22 & \underline{85.1} & 83.4 & 82.0 & 78.2 & 70.8 & 74.4 & 67.6 & 64.4 & 45.2 & 36.0 & 33.2 & 43.5 & 14.4 & 14.0 & 13.2 & 10.2 \\
 & MIAN & TKDE'23 & 15.5 & 22.1 & 28.5 & 29.0 & 11.0 & 14.3 & 16.2 & 15.0 & 8.6 & 9.6 & 10.9 & 10.7 & 7.0 & 7.3 & 7.0 & 7.7 \\
 & DHRL & TBD'24 & 10.6 & 38.1 &  86.5 & \textbf{91.0} & 10.1 & 65.5 & 83.5 &  \underline{86.5} & 10.7 & 39.3 & 68.6 & 72.1 & 8.2 & 8.3 & 39.9 &  45.5 \\
 & DHaPH & TKDE'24 & 78.9 & 84.9 & \underline{88.2} & 89.8 & 69.5 & 77.9 & \underline{83.9} & 86.0 & 51.5 & 64.9 & 72.9 & 80.7 & 42.3 & 49.2 & 60.0 & 70.9 \\
\multirow{-13}{*}{T2I} & RSHNL & Ours & \textbf{86.5} & \textbf{89.4} & \textbf{91.0} &  \underline{90.6} & \textbf{87.4} & \textbf{90.2} & \textbf{90.3} &  \textbf{90.3} & \textbf{82.6} & \textbf{87.4} & \textbf{89.1} & \textbf{88.8} & \textbf{79.7} & \textbf{86.8} & \textbf{86.7} & \textbf{85.3} 
\\ \hline
\end{tabular}
\label{tab:1}
\end{table*}

 % of the $t$-th epoch
\subsection{The Objective Function}
By combining the above losses, we can obtain the overall objective loss function as follows
\begin{equation}
\begin{aligned}
\mathcal{L}= 
\begin{cases}
\mathcal{L}_{p}+ \alpha\mathcal{L}_C,  & \text{if } t <  N_{w},  \\ \mathcal{L}_S+ \alpha\mathcal{L}_C  & \text{if }N_{w} \leq t < N_{m}.
\end{cases}
\end{aligned}
\label{eq:10}	
\end{equation}
where $\alpha$ is a hyper-parameter, $t$ is the current epoch, $N_{w}$ and $N_{m}$ are warm-up epoch and maximal epoch, respectively. The training process of RSHNL is shown in the appendix.

% \begin{algorithm}[tb]
% \caption{The pseudo-code of our RSHNL}
% \label{alg: algorithm}
% \textbf{Input}: Multi-modal data $\mathcal{D} =\left \{ \mathcal{D} _m \right \} _{m=1}^M$, $\tau$, $r$, $\alpha$, $\gamma$, hash functions $\left \{ \mathcal{H}^m (\cdot ,\Theta^m ) \right \} _{m=1}^M$, hash length $L$, batch size $n_b$, learning rate $l_r$, warm up epoch $N_{warmup}$, and maximal epoch $N_{epochs}$. 
% \newline
% \textbf{Output}: Model parameters $\Theta^m$.
% \begin{algorithmic}[1]
% \STATE Randomly generate $K$ real-value centers $C =\left \{ c_1,c_2,...,c_K \right \}$.
% \FOR{$iter = 1,2,\cdots, N_{epochs}$}
% \FOR{$step = 1,2,\cdots,\left\lfloor\frac{N}{n_b}\right\rfloor$}
% \STATE Randomly sample a mini-batch $\Bar{\mathcal{D}}$ from $\mathcal{D}$.
% \STATE Learn hash codes for $\Bar{\mathcal{D}}$ by hash functions $\left \{ \mathcal{H}^m (\cdot ,\Theta^m ) \right \} _{m=1}^M$.
% \STATE  Update hash centers $C$ by $tanh$ function and normalization.
% \STATE Calculate $\mathcal{L}$ shown in Equation 10.
% \STATE Update hash functions parameters $\left \{ \Theta^m  \right \} _{m=1}^M$ and hash centers $C$ by calculating their back-propagation gradients:\newline
% $C=C-lr\frac{\partial \mathcal{L} }{\partial C}$ \newline
% $\Theta^m=\Theta^m-lr\frac{\partial \mathcal{L} }{\partial \Theta^m},\; m=1,2,\cdots,M$
% \ENDFOR
% \ENDFOR
% \end{algorithmic}
% \end{algorithm}

\begin{table*}[]
\centering
\setlength{\tabcolsep}{1mm}
\caption{The MAP scores with different bit lengths on the INRIA-Websearch dataset under different noise rates.}
 \begin{tabular}{c|c|c|c|c|c|c|c|c|c|c|c|c|c|c|c|c|c|c}
 \hline
 &  & Noise & \multicolumn{4}{c|}{0.2} & \multicolumn{4}{c|}{0.4} & \multicolumn{4}{c|}{0.6} & \multicolumn{4}{c}{0.8} \\ \cline{3-19} 
 \multirow{-2}{*}{Task}& \multirow{-2}{*}{Method} & Ref. & 16 & 32 & 64 & 128 & 16 & 32 & 64 & 128 & 16 & 32 & 64 & 128 & 16 & 32 & 64 & 128 \\ \cline{1-19} 
 & DGCPN & AAAI'21 & 24.3 & 32.5 & 37.7 & 37.3 & 24.3 & 32.5 & 37.7 & 37.3 & \underline{24.3} & \underline{32.5} & \underline{37.7} & 37.3 &\underline{24.3} & \underline{32.5} & \underline{37.7} & 37.3 \\
 & CIRH & TKDE'22 & 14.6 & 21.5 & 26.5 & 30.8 & 14.6 & 21.5 & 26.5 & 30.8 & 14.6 & 21.5 & 26.5 & 30.8 & 14.6 & 21.5 & 26.5 & 30.8 \\
 & UCCH & TPAMI'23 & 18.9 & 26.1 & 29.5 & 34.3 & 18.9 & 26.1 & 29.5 & 34.3 & 18.9 & 26.1 & 29.5 & 34.3 & 18.9 & 26.1 & 29.5 & 34.3 \\
 & WASH & TKDE'23 & \underline{31.5} & \underline{38.0} & \underline{43.5} & \underline{46.2} & 26.2 & \underline{33.0} & \underline{38.5} & \underline{42.1} & 16.9 & 22.1 & 27.9 & 31.5 & 4.7 & 7.5 & 10.9 & 13.7  \\
 & HCCH & TMM'24 & 11.8 & 20.5 & 28.5 & 37.6 & 8.8 & 13.1 & 21.9 & 33.9 & 5.4 & 8.3 & 12.8 & 22.2 & 3.3 & 3.2 & 5.5 & 10.2  \\
 & DSCMH & AAAI'24 & 18.8 & 27.6 & 35.2 & 41.2 & 17.9 & 25.0 & 33.4 & 39.6 & 13.5 & 19.4 & 27.7 & 34.1 & 5.6 & 9.7 & 16.7 & 21.6  \\
 & HMAH & TMM'22 & 26.4 & 34.5 & 39.8 & 42.1 & 19.0 & 28.3 & 34.4 & 38.4 & 10.3 & 17.5 & 24.1 & 29.1 & 5.2 & 8.3 & 13.4 & 18.2 \\
 & CMMQ & CVPR'22 & 31.1 & 35.5 & 38.3 & 39.6 & \underline{27.8} & 32.1 & 34.4 & 35.8 & 17.0 & 21.2 & 27.9 & 30.4 & 4.1 & 3.9 & 11.6 & 9.8 \\
 & MIAN & TKDE'23 & 2.7 & 2.7 & 2.2 & 2.8 & 2.7 & 1.7 & 2.8 & 2.8 & 2.8 & 2.4 & 1.7 & 1.7 & 2.8 & 2.3 & 1.8 & 1.4 \\
 & DHRL & TBD'24 & 2.8 & 4.2 & 33.6 & 33.8 & 2.6 & 2.8 & 23.8 & 24.3 & 2.7 & 2.7 & 12.6 & 8.3 & 2.8 & 3.0 & 4.9 & 6.2 \\
 & DHaPH & TKDE'24 & 24.0 & 32.8 & 39.5 & 44.3 & 22.3 & 29.3 & 37.0 & 42.0 & 19.9 & 28.1 & 34.7 & \underline{39.8} & 19.5 & 26.1 & 33.6 & \underline{38.5} \\
\multirow{-15}{*}{I2T} & RSHNL & Ours & \textbf{39.3} & \textbf{48.0} & \textbf{51.9} & \textbf{52.4} & \textbf{37.9} & \textbf{45.8} & \textbf{50.3} & \textbf{51.6} & \textbf{31.2} & \textbf{38.3} & \textbf{47.6} & \textbf{49.0} & \textbf{28.3} & \textbf{38.2} & \textbf{41.8} & \textbf{42.9} \\ \hline
 & DGCPN & AAAI'21 & 22.9 & 32.0 & 37.5 & 36.9 & 22.9 & 32.0 & 37.5 & 36.9 & \underline{22.9} & \underline{32.0} & \underline{37.5} & 36.9 & \underline{22.9} & \underline{32.0} & \underline{37.5} & 36.9 \\
 & CIRH & TKDE'22 & 14.2 & 21.2 & 26.6 & 31.2 & 14.2 & 21.2 & 26.6 & 31.2 & 14.2 & 21.2 & 26.6 & 31.2 & 14.2 & 21.2 & 26.6 & 31.2 \\
 & UCCH & TPAMI'23 & 17.4 & 25.3 & 29.5 & 34.9 & 17.4 & 25.3 & 29.5 & 34.9 & 17.4 & 25.3 & 29.5 & 34.9 & 17.4 & 25.3 & 29.5 & 34.9 \\
 & WASH & TKDE'23 & 30.8 & \underline{38.4} & \underline{44.8} & \underline{47.8} & 25.1 & \underline{32.7} & \underline{39.3} & \underline{43.3} & 15.8 & 21.7 & 27.8 & 31.9 & 4.2 & 7.2 & 10.6 & 13.3  \\
 & HCCH & TMM'24 & 12.7 & 23.6 & 35.0 & 42.5 & 9.4 & 17.0 & 29.4 & 39.0 & 5.9 & 10.8 & 19.7 & 29.3 & 3.0 & 3.6 & 7.1 & 13.1  \\
 & DSCMH & AAAI'24 & 19.1 & 27.0 & 32.8 & 37.6 & 18.5 & 24.6 & 31.1 & 35.7 & 13.9 & 18.5 & 24.4 & 30.7 & 5.6 & 8.7 & 15.6 & 19.0  \\
 & HMAH & TMM'22 & 25.0 & 34.5 & 40.8 & 43.9 & 18.2 & 28.3 & 34.9 & 39.7 & 9.9 & 17.1 & 24.4 & 29.4 & 4.7 & 7.9 & 13.1 & 17.7 \\
 & CMMQ & CVPR'22 & \underline{31.4} & 36.6 & 38.5 & 39.0 & \underline{27.3} & 32.3 & 34.5 & 34.1 & 16.2 & 20.6 & 27.3 & 30.4 & 4.3 & 3.4 & 10.7 & 8.5 \\
 & MIAN & TKDE'23 & 1.2 & 1.2 & 1.2 & 1.2 & 1.2 & 1.2 & 1.2 & 1.2 & 1.2 & 1.2 & 1.2 & 1.2 & 1.2 & 1.2 & 1.2 & 1.2 \\
 & DHRL & TBD'24 & 2.7 & 4.2 & 32.3 & 33.7 & 2.7 & 2.6 & 23.5 & 24.0 & 2.8 & 2.6 & 13.2 & 7.7 & 2.7 & 2.9 & 4.8 & 7.1 \\
 & DHaPH & TKDE'24 & 22.4 & 32.8 & 40.7 & 45.7 & 21.0 & 29.3 & 37.6 & 43.1 & 18.6 & 28.4 & 35.4 & \underline{40.8} & 18.6 & 25.6 & 33.8 & \underline{39.5} \\
\multirow{-13}{*}{T2I} & RSHNL & Ours & \textbf{38.2} & \textbf{47.9} & \textbf{52.1} & \textbf{53.3} & \textbf{36.2} & \textbf{45.9} & \textbf{50.3} & \textbf{52.3} & \textbf{30.2} & \textbf{37.7} & \textbf{47.9} & \textbf{49.8} & \textbf{27.1} & \textbf{38.0} & \textbf{42.0} & \textbf{42.9} \\ \hline
\end{tabular}
\label{tab:2}
\end{table*}

\begin{table*}[]
\centering
\setlength{\tabcolsep}{1mm}
\caption{The MAP scores with different bit lengths on the XMediaNet dataset under different noise rates.}
\begin{tabular}{c|c|c|c|c|c|c|c|c|c|c|c|c|c|c|c|c|c|c}
\hline
 &  & Noise & \multicolumn{4}{c|}{0.2} & \multicolumn{4}{c|}{0.4} & \multicolumn{4}{c|}{0.6} & \multicolumn{4}{c}{0.8} \\ \cline{3-19} 
 \multirow{-2}{*}{Task}& \multirow{-2}{*}{Method} & Ref. & 16 & 32 & 64 & 128 & 16 & 32 & 64 & 128 & 16 & 32 & 64 & 128 & 16 & 32 & 64 & 128 \\ \cline{1-19}  
 & DGCPN & AAAI'21 & / & / & / & / & / & / & / & / & / & / & / & / & / & / & / & / \\
 & CIRH & TKDE'22 & / & / & / & / & / & / & / & / & / & / & / & / & / & / & / & / \\
 & UCCH & TPAMI'23 & 9.4 & 13.5 & 17.1 & 19.7 & \underline{9.4} & 13.5 & 17.1 & 19.7 & \underline{9.4} & 13.5 & 17.1 & 19.7 & \underline{9.4} & 13.5 & 17.1 & 19.7 \\
 & WASH & TKDE'23 & 8.0 & 15.1 & \underline{24.2} & \underline{34.0} & 7.0 & 13.1 & 21.2 & \underline{30.6} & 4.9 & 8.5 & 14.4 & 21.8 & 2.0 & 3.1 & 4.7 & 6.9  \\
 & HCCH & TMM'24 & 1.6 & 2.0 & 4.8 & 15.0 & 1.4 & 1.5 & 3.4 & 12.1 & 1.3 & 1.4 & 2.2 & 5.8 & 0.8 & 0.9 & 1.1 & 1.9 \\
 & DSCMH & AAAI'24 & 5.2 & 9.9 & 18.1 & 28.7 & 4.7 & 8.5 & 14.9 & 24.7 & 3.4 & 5.4 & 10.3 & 18.5 & 1.8 & 2.3 & 3.9 & 6.6  \\
 & HMAH & TMM'22 & 2.9 & 3.6 & 5.9 & 10.9 & 1.3 & 1.5 & 1.9 & 3.9 & 1.0 & 1.0 & 1.3 & 2.2 & 1.0 & 1.1 & 1.3 & 1.7 \\
 & CMMQ & CVPR'22 & / & / & / & / & / & / & / & / & / & / & / & / & / & / & / & / \\
 & MIAN & TKDE'23 & 0.8 & 1.5 & 1.8 & 2.5 & 0.8 & 1.2 & 1.3 & 1.8 & 0.7 & 0.9 & 1.0 & 1.2 & 0.7 & 0.8 & 0.8 & 0.9 \\
 & DHRL & TBD'24 & 0.7 & 0.7 & 0.9 & 15.2 & 0.7 & 0.7 & 1.0 & 13.2 & 0.7 & 0.7 & 2.7 & 6.3 & 0.7 & 0.7 & 0.7 & 1.3 \\
 & DHaPH & TKDE'24 & \underline{9.8} & \underline{15.9} & 23.4 & 28.5 & 9.1 & \underline{15.8} & \underline{22.6} & 27.6 & 9.2 & \underline{15.1} & \underline{21.9} & \underline{27.5} & 8.8 & \underline{15.2} & \underline{21.5} & \underline{27.3} \\
\multirow{-15}{*}{I2T} & RSHNL & Ours & \textbf{35.3} & \textbf{43.5} & \textbf{47.5} & \textbf{48.6} & \textbf{35.5} & \textbf{42.1} & \textbf{47.0} & \textbf{45.8} & \textbf{33.2} & \textbf{41.8} & \textbf{46.2} & \textbf{45.0} & \textbf{28.2} & \textbf{39.6} & \textbf{44.9} & {\color[HTML]{333333} \textbf{38.9}} \\ \hline
 & DGCPN & AAAI'21 & / & / & / & / & / & / & / & / & / & / & / & / & / & / & / & / \\
 & CIRH & TKDE'22 & / & / & / & / & / & / & / & / & / & / & / & / & / & / & / & / \\
 & UCCH & TPAMI'23 & 10.6 & 15.1 & 19.4 & 22.1 & \underline{10.6} & 15.1 & 19.4 & 22.1 & \underline{10.6} & 15.1 & 19.4 & 22.1 & \underline{10.6} & 15.1 & 19.4 & 22.1 \\
 & WASH & TKDE'23 & 10.3 & 17.3 & 25.8 & \underline{35.5} & 8.9 & 15.2 & 22.8 & \underline{32.2} & 6.3 & 10.2 & 15.9 & 23.4 & 2.4 & 3.6 & 5.4 & 7.8  \\
 & HCCH & TMM'24 & 1.8 & 1.5 & 2.1 & 9.0 & 1.4 & 1.2 & 1.7 & 12.1 & 1.5 & 1.1 & 1.3 & 1.8 & 0.8 & 0.9 & 1.0 & 1.2  \\
 & DSCMH & AAAI'24 & 4.1 & 8.4 & 16.5 & 27.5 & 3.6 & 7.0 & 13.5 & 23.8 & 2.5 & 4.4 & 9.3 & 17.5 & 1.4 & 2.0 & 3.4 & 5.9  \\
 & HMAH & TMM'22 & 3.5 & 4.4 & 6.6 & 12.1 & 1.6 & 1.8 & 2.3 & 4.4 & 1.1 & 1.2 & 1.5 & 2.5 & 1.0 & 1.2 & 1.4 & 1.8 \\
 & CMMQ & CVPR'22 & / & / & / & / & / & / & / & / & / & / & / & / & / & / & / & / \\
 & MIAN & TKDE'23 & 0.8 & 1.2 & 1.2 & 1.8 & 0.7 & 1.0 & 1.0 & 1.3 & 0.7 & 0.8 & 0.8 & 1.0 & 0.7 & 0.7 & 0.7 & 0.8 \\
 & DHRL & TBD'24 & 0.7 & 0.8 & 0.9 & 19.8 & 0.8 & 0.9 & 1 & 15.1 & 0.7 & 0.8 & 3.1 & 8.7 & 0.8 & 0.8 & 0.8 & 1.5 \\
 & DHaPH & TKDE'24 & \underline{11.0} & \underline{18.0} & \underline{26.3} & 32.6 & 10.5 & \underline{17.3} & \underline{25.3} & 31.6 & 10.3 & \underline{16.8} & \underline{24.8} & \underline{31.2} & 9.6 & \underline{17.1} & \underline{24.0} & \underline{30.7} \\
\multirow{-13}{*}{T2I} & RSHNL & Ours & \textbf{35.3} & \textbf{42.5} & \textbf{46.5} & \textbf{47.8} & \textbf{35.0} & \textbf{41.6} & \textbf{46.9} & \textbf{46.6} & \textbf{33.3} & \textbf{41.3} & \textbf{46.4} & \textbf{46.1} & \textbf{29.3} & \textbf{39.9} & \textbf{45.1} & \textbf{40.6} \\ \hline
\end{tabular}
\label{tab:3}
\end{table*}
\subsection{Theoretical Justification}
To show the robustness of the proposed RSHNL, we deeply analyze the impact of the weight $w_i$ on the loss $\mathcal{L}_S$. Our goal is to minimize $\mathcal{L}_S$ by updating the weight $w_i$ and network parameters $\left \{ \Theta^m \right \} _{m=1}^M$ alternatively, while making the other fixed. Given a fixed $\left \{ \Theta^m \right \} _{m=1}^M$, we can obtain the following optimal solution, i.e.,
\begin{equation}
\begin{aligned}
w_i^*&=\underset{w_i\in\left [ 0,1 \right ] }{\operatorname{argmin}}w_i\ell _i+\gamma (\frac{1}{2}{w_i}^2-w_i ) \\
&=\underset{w_i\in\left [ 0,1 \right ] }{\operatorname{argmin}}\frac{\gamma }{2} {w_i}^2+(\ell _i-\gamma )w_i.
\end{aligned}
\label{eq:11}	
\end{equation}
Since $w_i \in \left [ 0,1 \right]$, when $\ell_i - \gamma > 0$, the optimal solution $w_i^*$ is obviously 0. While $\ell_i - \gamma \le 0$, let the derivative of Eq.\ref{eq:11} be 0, we can obtain
\begin{equation}
\begin{aligned}
w_i^* = 1-\frac{\ell _i}{\gamma }.
\end{aligned}
\label{eq:12}	
\end{equation}
Clearly, since $\gamma \geq 0$ and $\ell_i \geq 0$, we can get $w_i \in [0,1]$ consistent with the original setting. In summary, we can get the solution as follows
\begin{equation}
\begin{aligned}
w_i^* = \max (0,1-\frac{\ell _i}{\gamma }).
\end{aligned}
\label{eq:13}	
\end{equation}

If the loss $\ell_i$ is too large (i.e., $\ell_i > \gamma$), we regard the corresponding sample pair as hard data with noise labels and assign a weight as $0$. When $\ell_i \leq \gamma$, the $i$-th pair with a large weight can be implicitly considered as easy. Otherwise, one with a small weight can be regarded as hard. In general, this strategy can not only distinguish clean sample pairs but also gradually train the hash model from easy to hard, embracing more robustness and generalization simultaneously. 
% The experimental analyses are shown in Fig.\ref{fig:3}

However, selecting a suitable learning pace parameter $\gamma$ is challenging. 
% As shown in Eq.\ref{eq:11}, 
If $\gamma$ is too small, no sample pairs would be selected for training. And if $\gamma$ is too large, all pairs will participate in training. Clearly, it will cause our NSH to be unable to distinguish noise labels, thus reducing the retrieval performance. In Eq.\ref{eq:6}, since $v_i^m\in\left [ 0,1 \right ]$ and $r > 0$, we can get the minimum value of $\ell_i$ as
\begin{equation}
\begin{aligned}
\ell _i^{min}=&0.
\end{aligned}
\label{eq:14}	
\end{equation}
Similarly, we can obtain the maximum value of $\ell_i$ as
\begin{equation}
\begin{aligned}
\ell _i^{max}=\frac{M(r^2-r+1)}{r}.
\end{aligned}
\label{eq:15}	
\end{equation}
Hence, we can get $\gamma$ is bounded by $0< \gamma < \frac{M(r^2-r+1)}{r}$. 
To obtain a suitable $\gamma$ to distinguish clean pairs, we perform the sensitivity analysis in the appendix.

\section{Experiments}
\subsection{Dataset}
To verify the effectiveness of the proposed RSHNL, we conduct extensive experiments on four widely used datasets, i.e., XMedia \cite{Xmedia}, INRIA-Websearch \cite{INRIA-Websearch}, Wikipedia \cite{wiki}, and XMediaNet \cite{XmediaNet}. 

\subsection{Experiments Settings}
% For all modalities, we stack three fully-connected (FC) layers on the backbones to learn hash codes. Specifically, the first two FC layers own 4096 neurons and follow a ReLU activation function, while the last FC layer owns $L$ neurons and follows a $\operatorname{tanh}$ function and a $L2$ regularization. For all datasets, we set the batch size $n_b$, the maximal epoch $N_{epochs}$, the warm-up epoch $N_{warmup}$, the weight factor $r$,  the learning rate $l_r$, and the temperature parameter $\tau$ as $256$, $150$, $5$, $0.01$, $0.0001$, and $1$ respectively. For four datasets, we set the hyperparameter $\alpha$ as $0.10$, $0.30$, $0.05$, and $0.05$ respectively and set the learning pace parameter $\gamma$ as $3$, $5$, $8$, and $9$ respectively. 
To evaluate the performance of the proposed RSHNL and competitors, we conduct two common cross-modal retrieval tasks. I2T and T2I represent using images as queries to retrieve texts and using texts as queries to retrieve images, respectively. Similar to \cite{qin2023cross}, we adopt Mean Average Precision (MAP) to evaluate the retrieval performance of all methods, which is a widely used evaluating metric. To comprehensively evaluate the effectiveness, we set noisy labels as symmetric noise with different rates (i.e., 0.2, 0.4, 0.6, and 0.8). Besides, the bit lengths are configured to 16, 32, 64, and 128. Besides, all experiments are conducted on a single GeForce RTX3090Ti 24GB GPU and our RSHNL is implemented in PyTorch 1.12.0. 
% For more detail, the network architectures and the parameters used in this paper are provided in the appendix due to space limitations.
More details on implementation are provided in the appendix due to space limitations.

\begin{figure}[htp]
\centering
\subfigure[PR curves (I2T)]{\includegraphics[width=0.20\textwidth]{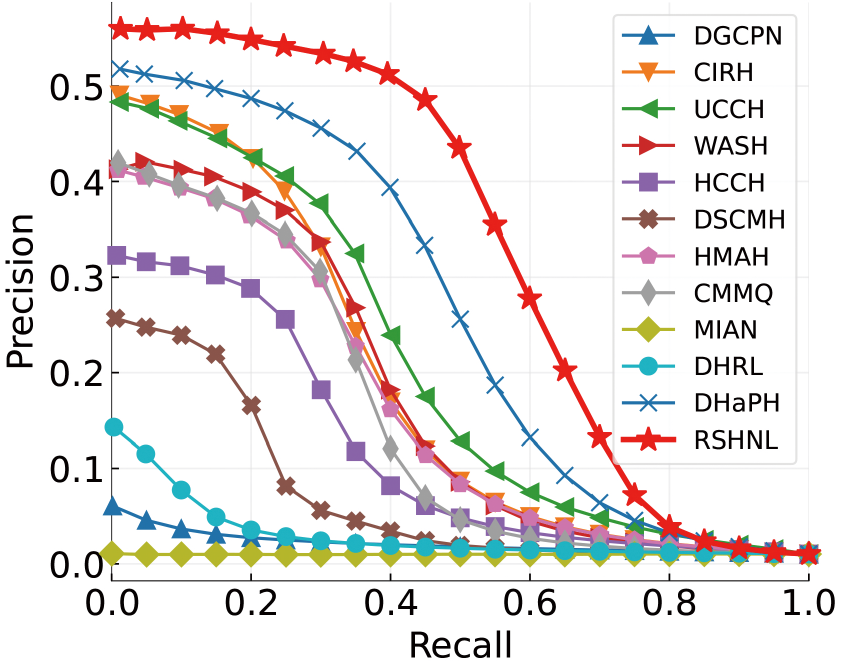}}
\subfigure[PR curves (T2I)]{\includegraphics[width=0.20\textwidth]{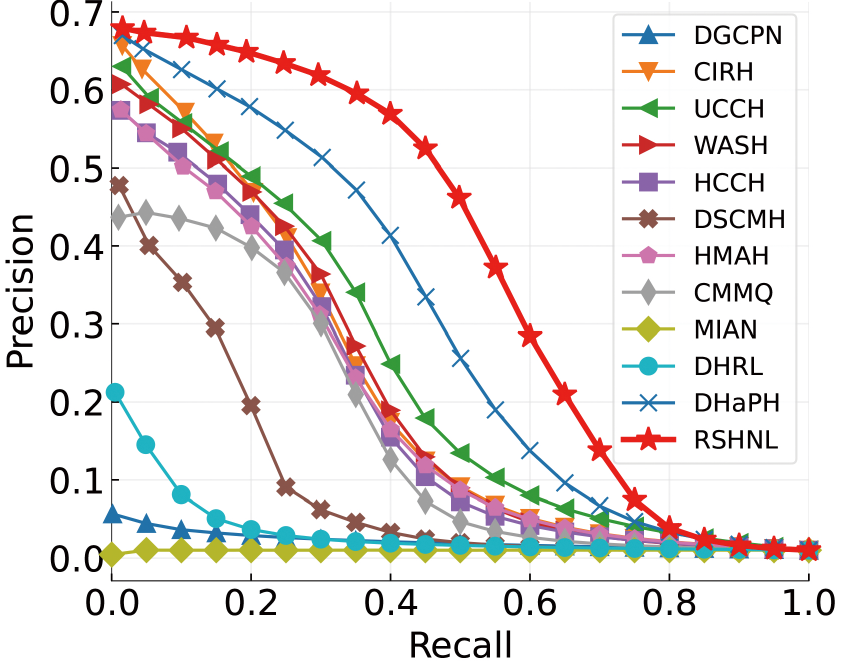}}
\subfigure[MAP curves (I2T)]{\includegraphics[width=0.20\textwidth]{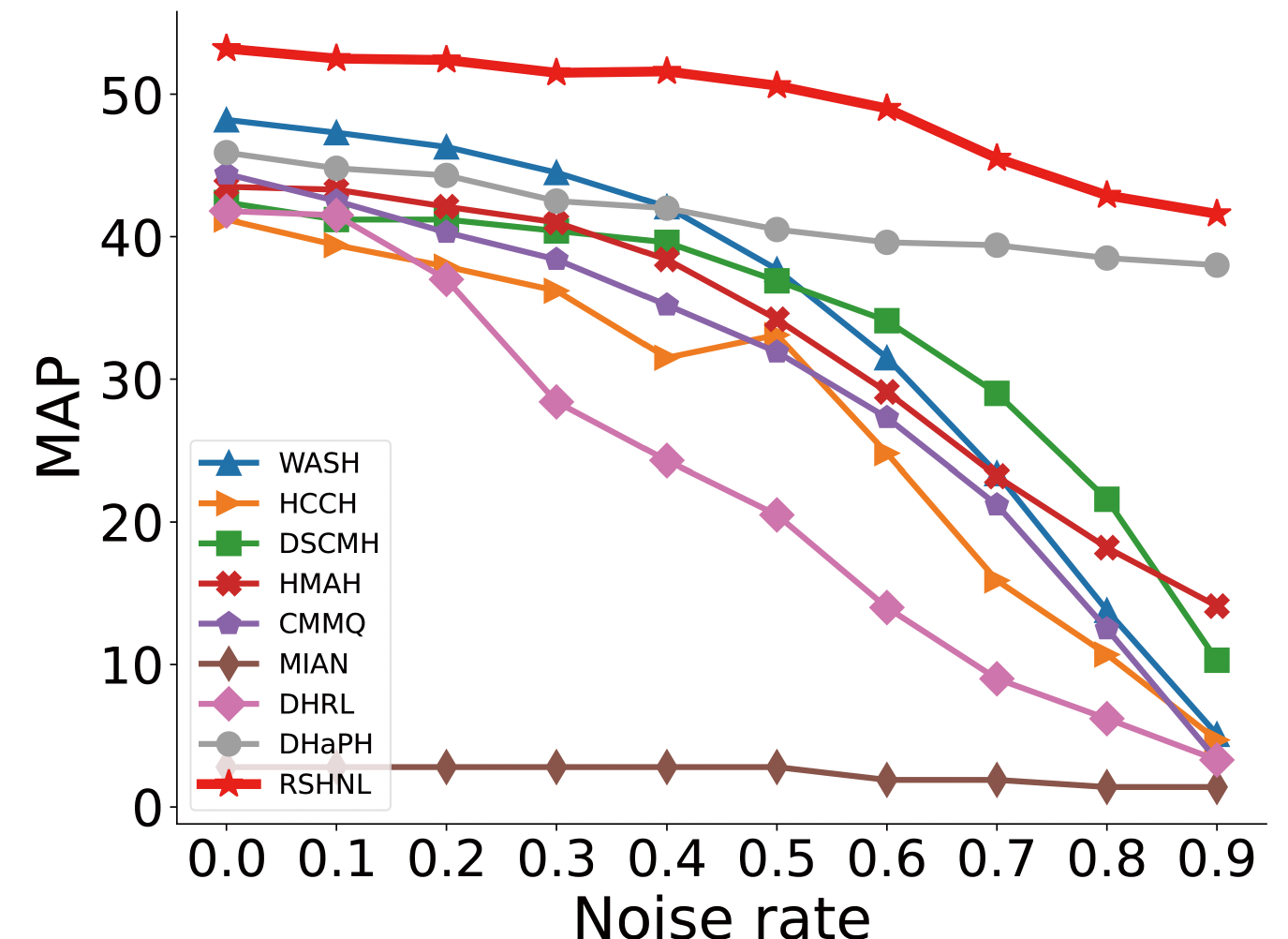}}
\subfigure[MAP curves (T2I)]{\includegraphics[width=0.20\textwidth]{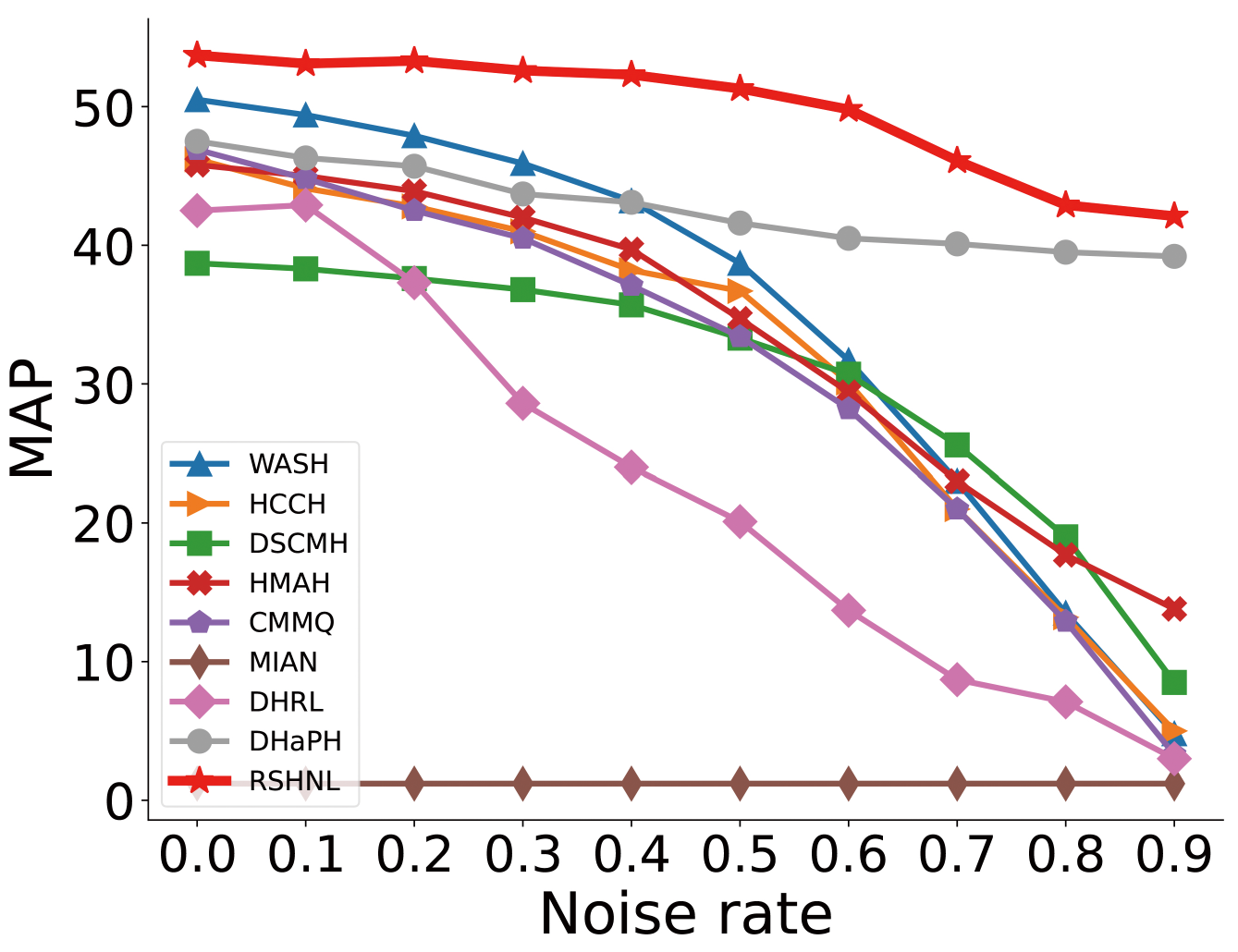}}
\caption{Experimental results with 128 bits on the INRIA-Websearch dataset under 0.6 noise rate.}
\label{fig:2}
\end{figure}

\subsection{Comparison methods}
To demonstrate the superiority of the proposed RSHNL, we compare RSHNL with 11 baselines, including deep unsupervised CMH methods (i.e., DGCPN \cite{DGCPN}, CIRH \cite{CIRH}, and UCCH \cite{UCCH}), shallow supervised CMH methods (i.e., WASH \cite{WASH}, HCCH \cite{sun2023hierarchical}, and DSCMH \cite{DSCMH}), and deep supervised CMH methods (i.e., HMAH \cite{HMAH}, CMMQ \cite{CMMQ}, MIAN \cite{MIAN}, DHRL \cite{DHRL}, and DHaPH \cite{DHaPH}). Among these, WASH, CMMQ, and DHRL are specifically designed to deal with the problem of noisy labels. DSCMH and DHaPH are the SPL-based hashing methods. For a fair comparison, we freeze the original backbones in the training process and report MAP scores on the testing set when MAP peaks on the validation set. For all experimental tables, the highest MAP scores are shown in \textbf{bold}, the second highest MAP scores are marked with \underline{underline}, and `/' denotes out-of-memory.

\subsection{Comparison with the State-of-the-Art}
The experimental results on three datasets are reported in \cref{tab:1,tab:2,tab:3}. The results on Wikipedia are given in the Appendix. Besides, we set the hash length as 128-bit on INRIA-Websearch, and then plot the precision-recall (PR) curves under 0.6 noise rate and the MAP curves under different noise rates in Fig.\ref{fig:2}. From these results, it can be observed:
\begin{itemize}
\item The retrieval performance of most methods improves with bit lengths increase because long hash codes contain more discriminative information. Besides, the performance of a few methods (such as CMMQ and RSHNL) degrades with bit lengths increase. This may be because it is difficult to resist the interference of noisy labels for these methods, resulting in more noise information being incorporated into long hash codes.
\item The retrieval performance of all supervised CMH methods is affected by noisy labels. As the noise rate increases, the CMH model is more likely to be misled, thereby resulting in a rapid drop in performance. Since unsupervised CMH methods do not need to exploit the label information, their performance is not affected by noisy labels at all.
\item Most methods show worse performance or even fail on the INRIA-Websearch and XMediaNet datasets because more categories significantly increase the difficulty of learning discriminative hash codes from noisy labels.
\item From PR curves, our RSHNL outperforms other baselines, which is consistent with what MAP demonstrates. According to MAP curves, the performance of almost all supervised CMH methods degrades as the noise rate increases. Thanks to the noise recognition capability of SPL, our RSHNL maintains stable and superior performance. Overall, the comprehensive performance of RSHNL outperforms all baselines. 
\end{itemize}

% Especially in the case of high noise rates, RSHNL shows fabulous noise resistance, demonstrating its effectiveness. 

% \begin{figure}[htp]
% \centering
% % \captionsetup[subfloat]{labelsep=none,format=plain,labelformat=empty}
% \subfigure[I2T]{
% \includegraphics[scale=0.18]{figures/noise_rate_0.1_0.9_I2T.pdf}}
% \subfigure[T2I]{
% \includegraphics[scale=0.18]{figures/noise_rate_0.1_0.9_T2I.pdf}}
% \caption{The MAP scores with 128 bits on the INRIA-Websearch dataset under different noise rates.}
% \label{fig:5}
% \end{figure}
% \begin{figure}[h]
% \centering
% \captionsetup[subfloat]{labelsep=none,format=plain,labelformat=empty}
% \subfloat[I2T]{
% \includegraphics[scale=0.25]{figures/wiki_old_I2TPR.pdf}}
% \subfloat[T2I]{
% \includegraphics[scale=0.25]{figures/wiki_old_T2IPR.pdf}}
%   \\ \text{(a)Wikipedia} \\
%   \subfloat[I2T]{
% 		\includegraphics[scale=0.25]{figures/INRIA-Websearch_I2TPR.pdf}}
%     \subfloat[T2I]{
% 		\includegraphics[scale=0.25]{figures/INRIA-Websearch_T2IPR.pdf}}
%   \\ \text{(b)INRIA-Websearch}
%   \caption{}
%   \label{fig: pr}
% \end{figure}

% \begin{figure}[htb]
% \centering
% \setlength{\abovecaptionskip}{0.cm}
% \subfigure[Different $\alpha$]{\includegraphics[width=0.23\textwidth]{figures/alpha_INRIA-Websearch.pdf}}
% \subfigure[Different $\gamma$]{\includegraphics[width=0.23\textwidth]{figures/gamma_INRIA-Websearch.pdf}}
% \caption{Parameter sensitivity analysis with 128 bits on INRIA-Wesearch dataset under 0.2 and 0.6 noise rates.}
% \label{fig:3}
% \end{figure}

\begin{figure}[htp]
\centering
% \captionsetup[subfloat]{labelsep=none,format=plain,labelformat=empty}
 \subfigure[Epochs=5]{
\includegraphics[width=0.20\textwidth]{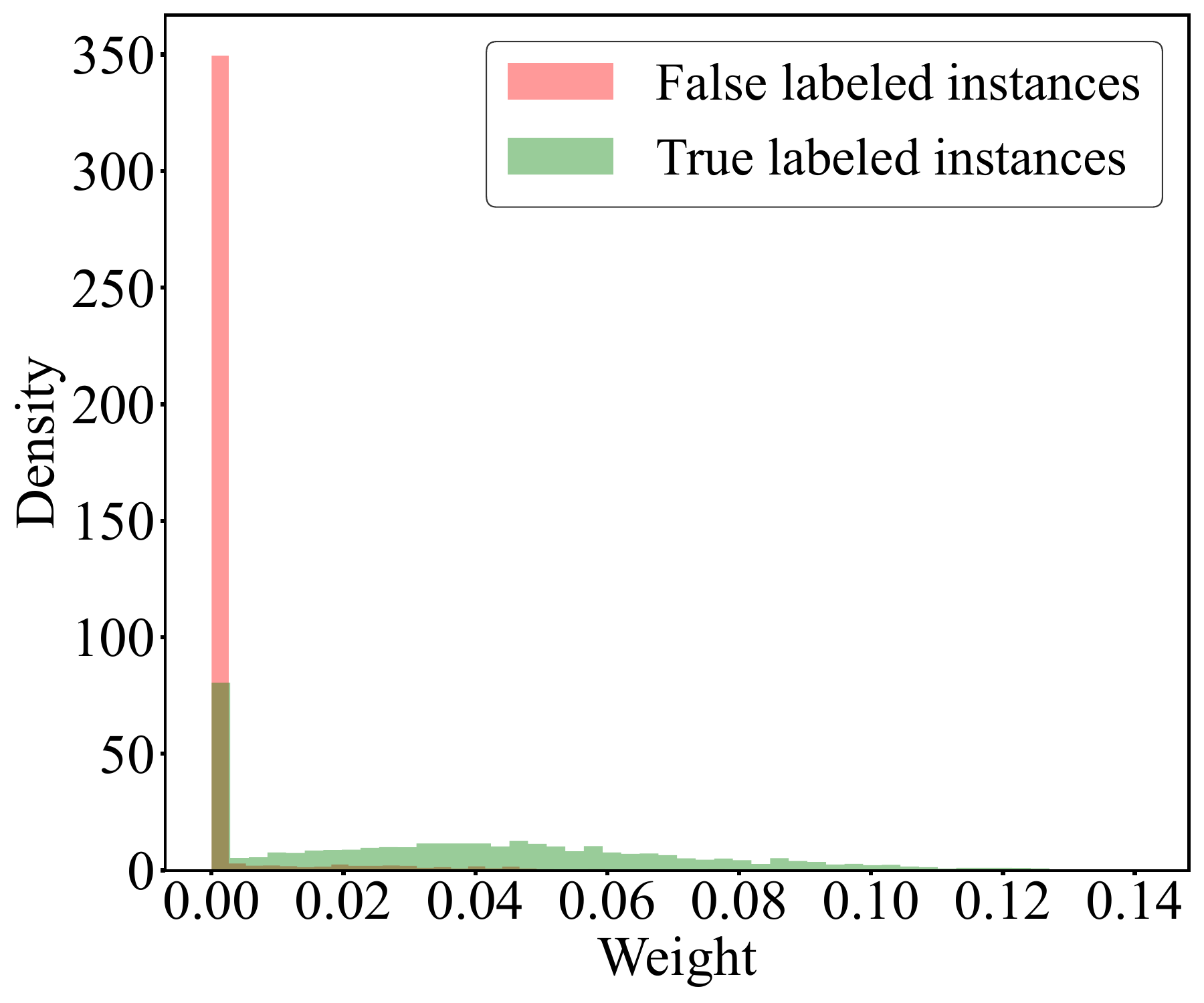}}
\subfigure[Epochs=100]{
\includegraphics[width=0.20\textwidth]{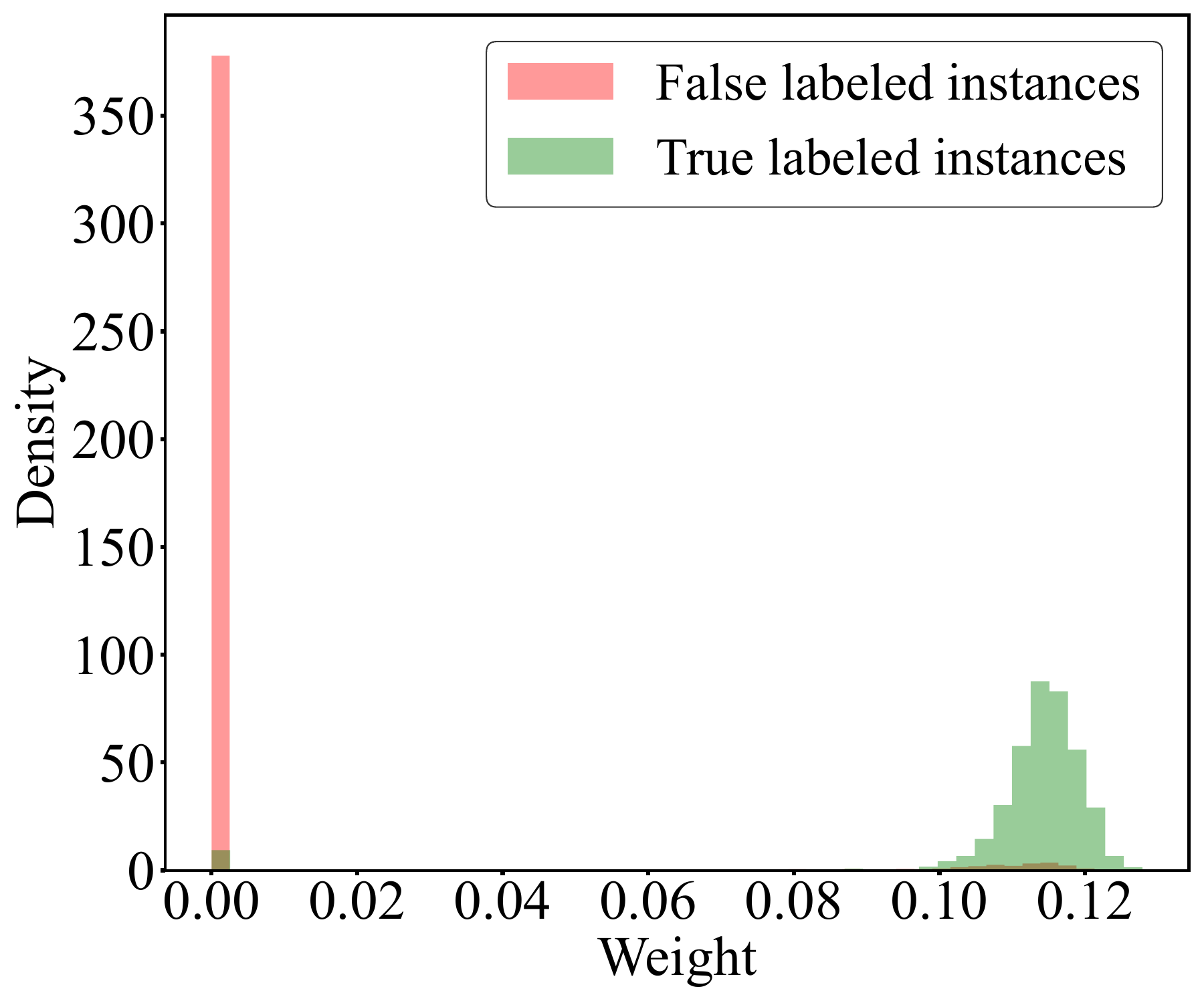}}
\caption{The density versus the weight of all instances with 128 bits and 0.6 noise rate.}
\label{fig:3}
\end{figure}

\subsection{Self-paced Analysis}
To study the self-paced behavior of our RSHNL with 128 bits and 0.6 noise rate, we plot the density versus the weight of each instance from different epochs on INRIA-Websearch. From Fig.\ref{fig:3}, we can observe that: 1) At the beginning, RSHNL first assigns zero weight to hard instances and regards them as noisy instances, thereby separating multi-modal data into a clean or noisy subset. 2) As the training progresses, RSHNL gradually learns with all clean instances from easy to hard until all instances become easy.

\subsection{Ablation Study}
To show the effectiveness of the proposed components, we conduct ablation experiments with 128 bits on the two datasets compared with three variants. Specifically, RSHNL-1, RSHNL-2, RSHNL-3 represent removing the warm-up training, removing the loss $\mathcal{L}_C$, and removing SPL of $\mathcal{L}_S$, respectively. To be fair, all variants adopt the same parameters as RSHNL. As shown in Tab.\ref{tab:4}, we report their average MAP scores on I2T and T2I tasks. From these results, RSHNL shows the best retrieval performance, which means that all components are crucial for RSHNL.

% 1) removing warming up will result in all instances not participating in NSH, so RSHNL-1 has the same performance under different noise rates. 2) removing SPL in $\mathcal{L}_S$ will significantly affect performance, demonstrating its effectiveness in learning with noisy labels. 3) in summary, removing any component of RSHNL will degrade the performance, demonstrating the benefits of each component. 

\begin{table}[htp]
\centering
\small
\caption{Ablation study with 128 bits.}
\setlength\tabcolsep{2pt}
\begin{tabular}{c|c|c|c|c|c|c|c|c}
\hline
Dataset & \multicolumn{4}{c|}{XMedia} & \multicolumn{4}{c}{INRIA-Websearch} \\ \hline
Noise & 0.2 & 0.4 & 0.6 & 0.8 & 0.2 & 0.4 & 0.6 & 0.8 \\ \hline
RSHNL-1 & 84.3 & 84.3 & 84.3 & 84.3 & 38.6 & 38.6 & 38.6 & 38.6 \\
RSHNL-2 & 90.1 & 85.6 & 72.1 & 25.4 & 50.1 & 45.6 & 20.2 & 8.7 \\
RSHNL-3 & 84.8 & 76.1 & 63.0 & 29.6 & 44.4 & 31.6 & 25.6 & 20.2 \\
RSHNL & \textbf{90.7} & \textbf{90.0} & \textbf{88.6} & \textbf{85.0} & \textbf{52.9} & \textbf{51.9} & \textbf{49.4} & \textbf{42.9 }\\ \hline
\end{tabular}
\label{tab:4}
\end{table}

\begin{figure}[htp]
\centering
\setlength{\abovecaptionskip}{0.cm}
\includegraphics[width=0.40\textwidth]{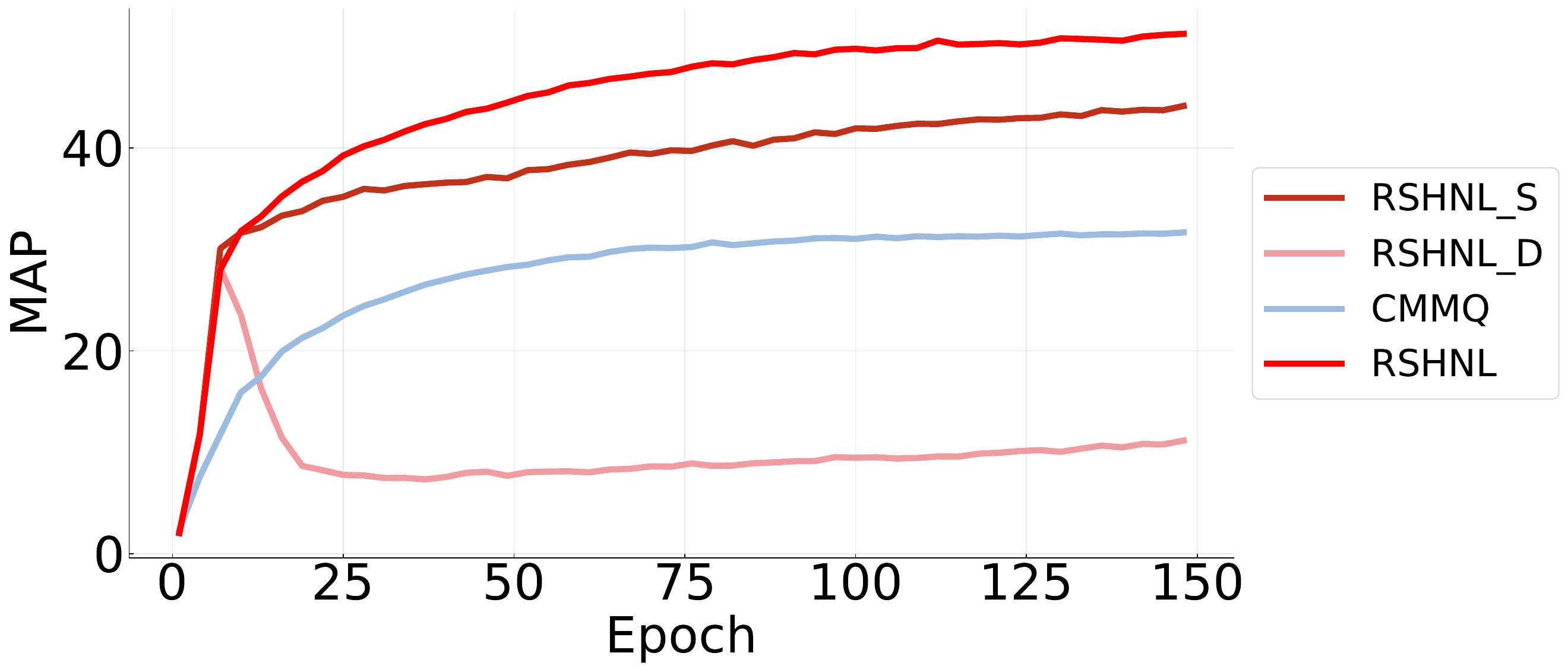}
\caption{The average MAP scores versus epochs.}
\label{fig:4}
\end{figure}

\subsection{Robustness Analysis}
To intuitively show the robustness of our RSHNL, on INRIA-Websearch under 0.6 noise, we compare it with CMMQ and two variants. Specifically, RSHNL-S represents removing the progressive learning mechanism by setting all weights greater than 0 to 1. RSHNL-D allows all instances to participate in the training by setting $\gamma$ as a value (e.g., 200) greater than $\ell_i^{max}$. Then, we plot the average MAP scores of I2T and T2I tasks with 128 bits. From Fig.\ref{fig:4}, we can observe that: 1) RSHNL-D overfits the noise, which indicates that the ability to distinguish noise is crucial. 2) Although CMMQ and RSHNL-S can prevent the overfitting problem, their retrieval performance is still lower than our method, which means our NSH could effectively improve the discrimination of hash codes by learning from easy to hard.

\section{Conclusion} 
In this paper, we propose a new cognitive cross-modal hashing approach (i.e., RSHNL) with noisy labels, which contains three parts, i.e., CHL, CAL, and NSH. Specifically, CHL maximizes the consistency of multi-modal data to alleviate the semantic gap. CAL learns a unified hash representation for each class as a center and encourages hash codes with the same category to be close to the corresponding hash centers. NSH presents a dynamic hardness measurement strategy that dynamically estimates the learning difficulty for each pair and distinguishes the noisy labels while facilitating learning hash codes from easy to hard for clean pairs. Extensive experiments show that RSHNL outperforms 11 state-of-the-art CMH methods under noisy labels.

\section{Acknowledgments}
This work is supported by the National Natural Science Foundation of China (Grant No. 62372315), the Sichuan Science and Technology Program (Grant No. 2024NSFTD0049, 2024ZDZX0004, 2024YFHZ0144, 2024YFHZ0089, MZGC20240057), and the Mianyang Science and Technology Program (Grant No. 2023ZYDF091, 2023ZYDF003).

% In this work, we propose a new Robust Self-paced Hashing with Noisy Labels (RSHNL), which can mitigate the negative impact of noisy labels and robustly learn discriminative binary codes. Specifically, RSHNL first proposes Contrastive Hashing Learning (CHL) to alleviate the cross-modal gap by maximizing the semantic consistency of multimodal pairs. Secondly, RSHNL proposes center aggregation learning (CAL) to learn shared aggregation centers to compact hash codes from the same category, thereby maintaining the similarity of intra-class binary codes. Next, RSHNL proposes a Noise-tolerance Self-paced Hashing (NSH) mechanism that can automatically distinguish noisy instances and select clean instances for learning from easy to `hard', thus embracing better robustness of the model. Lastly, extensive experiments are conducted on four widely-used datasets to demonstrate that our proposed RSHNL outperforms the state-of-the-art in both performance and robustness.
\bigskip
\noindent 

\bibliography{aaai25}

\end{document}